\documentclass[10pt,twocolumn,letterpaper]{article}

\usepackage{cvpr}
\usepackage{times}
\usepackage{epsfig}
\usepackage{graphicx}
\usepackage{amsmath}
\usepackage{amssymb}
\usepackage{subfigure}
\usepackage{xcolor}
\usepackage{epstopdf}

\usepackage[pagebackref=true,breaklinks=true,letterpaper=true,colorlinks,bookmarks=false]{hyperref}

\cvprfinalcopy 

\def\cvprPaperID{4035} 

\ifcvprfinal\fi

\newcommand{\U}{\mathbf{U}}
\newcommand{\V}{\mathbf{V}}

\newcommand{\m}{\mathbf{m}}

\newcommand{\M}{\mathbf{M}}
\newcommand{\Sm}{\mathbf{S}}

\newcommand{\T}{\mathrm{T}}

\newcommand{\bb}{\mathbf{b}}
\newcommand{\B}{\mathbf{B}}
\newcommand{\W}{\mathbf{W}}
\newcommand{\rank}{\mathrm{rank}\;}
\newcommand{\A}{\mathbf{A}}
\newcommand{\tr}{\mathbf{t}}
\newcommand{\x}{\mathbf{x}}

\newcommand{\D}{\mathbf{D}}

\title{Uncalibrated Non-Rigid Factorisation by Independent Subspace Analysis
}

\author{Sami S. Brandt\\
IT University of Copenhagen,\\ Copenhagen, Denmark\\
\and
Hanno Ackermann and Stella Grasshof\\
Leibniz Universit\"at Hannover,\\ Hannover, Germany\\
}

\begin{document}
\maketitle

\begin{abstract}
We propose a general, prior-free approach for the uncalibrated non-rigid structure-from-motion problem for modelling and analysis of non-rigid objects such as human faces. The word general refers to an approach that recovers the non-rigid affine structure and motion from 2D point correspondences by assuming that (1) the non-rigid shapes are generated by a linear combination of rigid 3D basis shapes, (2) that the non-rigid shapes are affine in nature, i.e., they can be modelled as deviations from the mean, rigid shape, (3) and that the basis shapes are statistically independent. In contrast to the majority of existing works, no prior information is assumed for the structure and motion apart from the assumption the that underlying basis shapes are statistically independent. The independent 3D shape bases are recovered by independent subspace analysis (ISA). Likewise, in contrast to the most previous approaches, no calibration information is assumed for affine cameras; the reconstruction is solved up to a global affine ambiguity that makes our approach simple but efficient. In the experiments, we evaluated the method with several standard data sets including a real face expression data set of 7200 faces with 2D point correspondences and unknown 3D structure and motion for which we obtained promising results.
\end{abstract}


\section{INTRODUCTION}

The estimation of structure and motion from image streams is a fundamental problem in computer vision. As an extension to the regular structure-from-motion (SFM) problem, the non-rigid structure-from-motion (NRSFM) problem takes the non-rigidity of the object in consideration in the recovery of structure and motion. The NRSFM problem has received considerable attention during the last two decades and encouraging results have been obtained. 

\begin{figure}[bt]
\begin{center}
\includegraphics[width=0.35\textwidth]{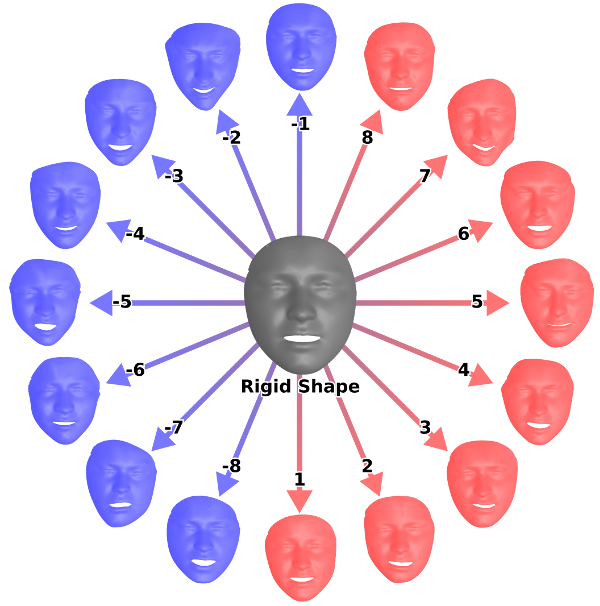}
\end{center}
\caption{
 We propose a 
method that infers the 3D reconstruction, basis shapes, and the underlying affine camera geometry from the 2D projections of a non-rigid object
by only assuming an uncalibrated affine camera and \emph{statistically independent} basis shapes.}
\label{fig:teaser}
\vspace{-5mm}
\end{figure}

The approaches for NRSFM can be categorised in several ways. From the algorithmic point of view, there are \emph{direct} and \emph{iterative} methods. Starting from the direct methods, 
the work of Bregler \etal\@ \cite{Bregler00} can be seen as the starting point for NRSFM research. They proposed an approach where the shape deformations are modelled as a linear combination of rigid shape basis that leads to a low-rank model; a heuristic 1D factorisation together with orthogonal constraints were used to recover the camera matrices. This pioneering work was thereafter succeeded by the work of Brand \etal\@ \cite{Brand01}, who used the heuristic of minimising deformations. Del Bue and Agapito applied additional constraints arising from a stereo rig \cite{DelBue04}.   
Xiao \etal\@ constrained the shape basis by assuming that each basis shape is visible unmixed in some frames \cite{Xiao06}. Hartley and Vidal proposed a solution for perspective non-rigid structure from motion problem by factoring a multifocal tensor \cite{Hartley08}. 

Regarding the iterative methods, one category is alternation-based methods, such as the trilinear method by Torresani \etal\@ \cite{Torresani01}, the bilinear methods by Paladini \etal\@ \cite{Paladini12} and Del Bue \etal\@ \cite{DelBue12} which include projections onto the metric manifold, and the 
method 
by Torresani \etal\@ \cite{Torresani08} which is based on Probabilistic PCA and Expectation Maximisation. 
Bundle adjustment has been applied, for instance, in \cite{Aanaes02b,DelBue06,Bartoli08}. Moreover, Bartoli \etal\@ \cite{Bartoli08} used a coarse-to-fine problem formulation to obtain a robust result. Various works have applied either statistical or physical priors to regularise the non-rigid structure from motion problem. These include 
priors such as rigidity \cite{DelBue06, Bartoli08}, smoothness of camera trajectories \cite{Gotardo11}, temporal smoothness \cite{Torresani08,Akhter2009:Trajectory}, deformation locality \cite{Brand01} and type \cite{DelBue12} have been used. 

When analysing, for instance, a set of face images without temporal order, temporal priors cannot be applied.
 An early prior free\footnote{By 'prior-free' 
 we refer to an approach that does not make an assumption, in the Bayesian sense, about the prior distribution of the basis shapes. 
} approach for uncalibrated non-rigid structure from motion was proposed by Brandt \etal\@ \cite{Brandt09} where the shape basis ambiguity was approached by assuming statistical independence between the basis shapes that led to independent subspace analysis (ISA).  
 Dai \etal's solution \cite{Dai12} for prior-free non-rigid structure-from-motion was built upon the observation by Akhter \etal\@ \cite{Akhter09}, namely that even though there is an unresolved ambiguity for shape basis with the standard orthogonality constraints, the 3D shape can be recovered without an ambiguity. Kong and Lucey \cite{Kong16} proposed a prior-free approach where the non-rigid shape is modelled as a compressible basis instead of a low-rank basis. 
%
 %
In applications such as facial expression analysis, it is also valuable to reconstruct the underlying shape basis and decompose the expressions onto it. This is a drawback for the approach of Dai \etal \cite{Dai12} where the shape basis is only implicitly present and ambiguous. Likewise, Kong and Lucey's \cite{Kong16} approach does not estimate a shape basis but a compressible feature basis.  


In this paper, 
we propose a prior-free, non-rigid structure from motion algorithm based on independent subspace analysis.  
The assumption hence is that the underlying shape bases live in \emph{statistically independent}\footnote{Two random variables $\mathbf{X}$, $\mathbf{Y}$ are statistically independent iff their joint probability density factorises, i.e., $p(\mathbf{x},\mathbf{y}) = p(\mathbf{x})p(\mathbf{y})$.} subspaces. Statistical independence should not be confused with linear independence used, for instance, by Xiao \etal\@ \cite{Xiao06} for selecting the shape basis. Remarkably, the statistically independent subspaces, and hence the basis shapes, can be recovered in an uncalibrated, affine setting, thus no calibration information, neither intrinsic nor extrinsic, of the affine cameras is required to infer the basis shapes (cf.~Fig.~\ref{fig:teaser}). This is a \emph{major simplification} of the non-rigid structure from motion problem. Furthermore, in contrast to the method by Brandt \etal\@ \cite{Brandt09}, the proposed method does not require an exhaustive search over  one-dimensional subspace permutations which constitutes an NP-hard problem.

{\bf The contributions} of this work are as follows. (1) We propose a straightforward, priorless, direct method for non-rigid structure from motion by assuming statistical independence of the basis shapes in an uncalibrated setting. 
(2) In contrast to many other non-rigid factorisation algorithms built upon the seminal algorithm of Bregler \etal\@ \cite{Bregler00}, our construction is based on the fact that all the shape bases are projected onto the image plane by a shared camera matrix. 
Moreover, we assume that the affine camera matrix will be solely defined by the mean, rigid shape -- this is consistent with Independent Subspace Analysis since it has been shown in  \cite{Hyvarinen02} that this is equivalent to analyse the original or mean corrected observations while the structure of the latter setting is simpler. 
(3) To recover the shape basis we suggest two alternative ISA algorithms built upon mutual information minimisation: FastISA proposed in \cite{Hyvarinen06} and FastICA \cite{Hyvarinen97} equipped with our component pooling. The algorithms do not require to exhaustively determine permutations of one-dimensional shape components in contrast to the method in \cite{Brandt09}. (4) To recover the block-formed motion matrix
after the ISA step, we propose an algebraic, iteratively re-weighted least squares method where only subspace affinities and the shape mixing coefficients are left to be estimated. 
(5) We propose a non-linear refinement method to obtain the final, statistically sound estimates. 


 %
    

%

\section{AFFINE NON-RIGID MODEL}

The standard non-rigid factorisation assumes that the non-rigid shape can be represented as a linear combination of the shape bases. That is, 3D points can be expressed as $\x^i_j=\sum_k \alpha_k^i \bb_{kj}$, where $\alpha_k^i$ is a scalar. 
We assume that the model is affine, i.e.~centred around the rigid, mean shape. 
The 2D projection $\hat{\m}_j^i$ of a 3D point $\x^i_j$ hence is
\begin{equation}
  \hat{\m}_j^i= \M^i \x^i_j  +{\tr}^i = \M^i \left(\bb_{0j} + \sum_{k=1}^K  \alpha_k^i \bb_{kj}  \right) +{\tr}^i,
\end{equation}
where $\M^i$ is $2\times3$ projection matrix to the image $i$, $\tr^i$ is the corresponding translation vector, 
$\alpha_k^i$, $k=1,2,\ldots,K$ are the scalar coefficients, and  $\bb_{kj}$ contains the basis shapes; $k=0$ refers to the mean rigid shape. 
 
Assuming Gaussian noise, the maximum likelihood solution with respect to the parameters  ${\M^i,\tr^i,\alpha_k^i,\bb_{kj}}$, $i=1,\ldots,I$, $j=1,\ldots,J$, $k=1,\ldots,K$,
minimises the cost 
\begin{equation}
\sum_{i,j} \| \hat{\m}_j^i-\m_j^i \|^2 =\sum_{i,j} \| \M^i (\bb_{0j} + \sum_k \alpha_k^i \bb_{kj})  +{\tr}^i - \m_j^i \|^2 \nonumber
\end{equation}
or equivalently 
\begin{equation}
  \| \W-\hat{\W} \|_\mathrm{Fro}^2, \label{eq:fro}
\end{equation}
where the translation corrected measurements $\m_j^i-\hat{\tr}^i$, $\hat{\tr}^i=\frac{1}{J} \sum_j \mathbf{m}_j^i$, are collected into the matrix $\W$, implying
\begin{equation}
{\W}\simeq\underset{\triangleq \M}{\underbrace{
    \begin{pmatrix}
       \M^1  &  \alpha_1^1 \M^1  & \cdots &  \alpha^1_K \M^1 \\
       \M^2 &  \alpha_1^2 \M^2 & \cdots &  \alpha^2_K \M^2 \\
      \vdots          & \vdots          & \ddots &  \vdots\\
       \M^I &  \alpha_1^I \M^I & \cdots &  \alpha^I_K \M^I \\          
    \end{pmatrix} }}\underset{\triangleq \B}{\underbrace{
    \begin{pmatrix}
      \B_0\\
      \B_1\\
      \vdots\\
      \B_K
    \end{pmatrix}}}, \label{eq:factorisedform}
\end{equation}
where $\B_k=\left( \bb_{k1} \ \bb_{k2}\ \cdots \ \bb_{kJ} \right)$ and $\B_0$ is the rigid shape. 
All the shape bases share the same inhomogeneous projection matrix $\M^i$ for image $i$. From
\eqref{eq:factorisedform} it follows that the noise free measurement
matrix has the rank constraint $R\triangleq\rank \hat{\W} \leq 3K+3$. 

The matrix minimising \eqref{eq:fro} with the rank constraint is
obtained by the singular value decomposition of $\W=\mathbf{U}
\mathbf{S} \mathbf{V}^\mathrm{T}$ by truncating the smallest
singular values, keeping the $3K+3$ largest, and removing the corresponding singular vectors. The truncated matrices being $\tilde{\mathbf{U}}$, $\tilde{\mathbf{S}}$ and $\tilde{\mathbf{V}}$ yields
\begin{equation} \label{eq:affineambiquity}
  \hat{\W} = \underset{\triangleq \tilde{\M}}{\underbrace{\left(\frac{1}{\sqrt{J}}\tilde{\mathbf{U}} \tilde{\mathbf{S}}\right)}} \underset{\triangleq \tilde{\B}}{\underbrace{\left(\sqrt{J}\tilde{\mathbf{V}}^\mathrm{T}\right)}}= \underset{\triangleq \hat{\M}}{\underbrace{\tilde{\M}\mathbf{A}}} \underset{\triangleq \hat{\B}}{\underbrace{\mathbf{A}^{-1}\tilde{\B}}}=\hat{\M}\hat{\B},
\end{equation}
where $\A$ is an unknown affine transformation. 
To find the estimates for the
non-rigid structure $\hat{\B}$ and motion matrix
$\hat{\M}$, we the need to find the linear transformation $\A$
that (1) separates the statistically independent shape subspaces and (2) recovers the block structure of the motion matrix. Our solution is described in the following section. 

\section{PROPOSED METHOD} 
 
This section describes the proposed method. It consists of the following steps: estimation of the rigid and non-rigid components (Sec.~\ref{sec:svd}), independent subspace analysis (Sec.~\ref{sec:isa}), block-form motion matrix recovery (Sec.~\ref{sec:block}), and non-linear refinement (Sec.~\ref{sec:refinement}).
 
\subsection{Factorisation} \label{sec:svd}

To facilitate ISA processing and for clarity, we divide the translation corrected measurement matrix into rigid and non-rigid part as follows. We first compute the nearest rigid affine reconstruction by the standard Tomasi--Kanade factorisation \cite{Tomasi92}  
that yields the rigid approximation
\begin{equation}
    \W_0 = \M_0 \B_0, 
    \label{eq:fac.tomasi}
\end{equation}
where the inhomogeneous projection matrices, up to an affine transform, 
are $\M_0=\frac{1}{\sqrt{J}}\U_0 \Sm_0$ and the mean rigid shape is $\mathbf{B}_0 
= \sqrt{J} \V_0^\T$. 
%
%
We then subtract the rigid component from the measurement matrix
\begin{equation}
    \Delta \W = \W - \W_0,
\end{equation}
and work with the non-rigid part $\Delta \W$.

 
 
Now, by using the remaining constraint $\rank \Delta \W \leq3K$ for the non-rigid part, we truncate all the singular values, and singular vectors, up to the $3K$ largest that yields 
\begin{equation} 
  \Delta \W \approx \Delta \tilde{\W} \equiv {\U}' {\Sm}' {\V}'^\T = {\M}' {\B}',\label{eq:svd}
\end{equation}
where ${\M}'=\frac{1}{\sqrt{J}}{\U}' {\Sm}'$ and  ${\mathbf{B}}'= \sqrt{J} {\V}'^\T$.

\subsection{Independent Subspace Analysis} \label{sec:isa}
 
As it is well known, the SVD step \eqref{eq:svd} does not generally yield the block structure to the motion matrix. 
By independent subspace analysis (ISA) we map the singular vectors into groups of three so that the groups will be as statistically independent to each other as possible. By ISA, we are searching for the orthogonal mixing matrix $\A_{\mathrm{ISA}}$ in the whitened space that maps the signals ${\B}_{\mathrm{ISA}}$ 
into the observed mixtures 
such that 
%
\begin{equation}
{\M}' {\B}'
= \underset{\triangleq \M_{\mathrm{ISA}}}{\underbrace{{\M}' \A_{\mathrm{ISA}}}} \underset{\triangleq \B_{\mathrm{ISA}}}{\underbrace{ \A_{\mathrm{ISA}}^\T {\B}'}} \equiv {\M}_{\mathrm{ISA}}  {\B}_{\mathrm{ISA}}.
 \end{equation}
Thereby, the rows in ${\B}_{\mathrm{ISA}}$ consist of statistically independent groups of three basis vectors. 

In this work, we use two ISA algorithms that yield an estimate for the orthonormal, subspace separation matrix $\A_{\mathrm{ISA}}^\T$. The first one (ISA1) is principally the FastISA algorithm \cite{Hyvarinen06} 
that has been developed from the FastICA \cite{Hyvarinen97} algorithm with the difference that the statistical independence of individual components is not assumed but instead between vectors residing in different subspaces. This is in analogy to assuming sparsity or group sparsity of multivariate signals.  However, the approach has been reported local hence relatively sensitive to intialisation. Moreover, the construction of FastISA is based on an \emph{ad hoc} probability density model that may compromise its statistical performance. To cope with the locality, we compute FastISA from multiple initialisations and take the estimate that maximises the likelihood of the solution with the density assumption. 

Our alternative ISA algorithm (ISA2) is the FastICA algorithm \cite{Hyvarinen97} followed by our own component pooling. ISA solution can namely be constructed by first estimating the one-dimensional ICA components, which are as independent as possible, and grouping them into subspaces. By ICA, the one-dimensional signal separation is computed by using the higher-order statistics of the basis vectors ${\mathbf{B}}'$. 
We may then additionally use the image population statistics to solve the component pooling problem. 
In more detail, we project the non-rigid measurement matrix $\Delta \mathbf{W}$ onto the orthogonal, $3K$-dimensional ICA basis $\mathbf{B}_\mathrm{ICA}=\A_{\mathrm{ICA}}^\T {\B'}$, and estimate the $3K \times 3K$ covariance matrix $\mathbf{C}=\frac{1}{J}\mathbf{B}_\mathrm{ICA} \Delta \mathbf{W}^\T \Delta \mathbf{W} \mathbf{B}_\mathrm{ICA}^\T-\frac{1}{4I^2J} \mathbf{B}_\mathrm{ICA} \Delta \mathbf{W}^\T \mathbf{1} \mathbf{1}^\T \Delta \mathbf{W} \mathbf{B}_\mathrm{ICA}^\T $ of these projections. For a statistically independent component pair, the covariance will vanish, \ie, the covariance matrix will show block diagonal structure, as soon as the components are correctly permuted. We thus estimate the ICA component permutation matrix $\mathbf{P}$, and further the orthogonal transformation $\A_{\mathrm{ISA}}^\T=\mathbf{P} \A_{\mathrm{ICA}}^\T$, by a greedy strategy: in analogy to using Givens rotations, we compute the sequence of optimal pairwise variable permutations that decrease the off-block-diagonal covariation in the covariance matrix.   

\subsection{Recovery of the Block Structure} \label{sec:block}
By a blind subspace separation method, the independent subspaces can be recovered only up to an unknown linear transform for each independent subspace, since the energy of the independent components cannot be recovered \cite{Hyvarinen02}. In other words, after ISA, we 
need to
estimate the $3\times3$ mapping $\D_k$ from the rigid shape coordinate system onto coordinate system of the independent subspace $k$.\footnote{In the calibrated case, $\D_k$ would be the $3\times3$ rotation $\mathbf{R}_k$ between the rigid and the non-rigid shape basis. 
Here, however, $\D_k$ is a general $3 \times 3$ matrix, constrained to unity norm to fix the arbitrary scale of the solution.} 
Let $\D$ be the block diagonal matrix containing all the $K$ subspace affinities in the respective blocks. We may then write 
\begin{equation}
{\M}_{\mathrm{ISA}}  {\B}_{\mathrm{ISA}} = \underset{\triangleq \hat{\M}}{\underbrace{\M_{\mathrm{ISA}} \D}} \underset{\triangleq \hat{\B}}{\underbrace{\D^{-1} \B_{\mathrm{ISA}}}} \equiv \hat{\M} \hat{\B}
\end{equation}
that also maps the motion matrix ${\M}_{\mathrm{ISA}}$ into the block-form matrix.
To compute an algebraic estimate for $\D$, we use the assumption that each shape basis component, including the rigid shape, share a common affine projection matrix to each view, 
and minimise
 \begin{equation}
     \min_{\D,\alpha}
     \sum_{i,k} \| \M_k^i \D_k - \alpha_k^i \M_0^i\|^2_\mathrm{Fro} \label{eq:alg}
\end{equation}
subject to $\| \D_k \|^2_\mathrm{Fro}=1$ for $k=1,2,\ldots,K$, where $\M_k^i$ is the $2 \times 3$ block of $\M_\mathrm{ISA}$, indexed by $k$ and $i$. The matrix $\M_0^i$ is the inhomogeneous affine projection matrix $i$ in $\M_0$ in \eqref{eq:fac.tomasi}. 
The estimate can be found by iteratively reweighted least squares as detailed in Appendix~\ref{app:IRLS}.

\subsection{Non-linear Refinement} \label{sec:refinement}

Since \eqref{eq:alg} is an algebraic criterion, we finally make a 
non-linear refinement to minimise the reprojection error, or 
\begin{equation}
    \min_{\D,\alpha}
    \| \W - \M_0^\alpha \D^{-1} \B_\mathrm{ISA} \|_{\mathrm{Fro}}^2
\end{equation}
where $\M_0^\alpha$ is defined by the matrix $M_0$ repeated $K$ times and the scalar weights $\alpha_k^i$ multiplied to the corresponding $2 \times 3$ blocks. 
Using the fact that the rows of $\frac{1}{\sqrt{J}}\B_\mathrm{ISA}$ are orthonormal,
\begin{equation}
\begin{split}
   \| \W &- \M_0^\alpha \D^{-1} \B_\mathrm{ISA} \|_{\mathrm{Fro}}^2 \\
   &=  \left \| \frac{1}{J} \W \B_\mathrm{ISA}^\T \B_\mathrm{ISA} - \M_0^\alpha \D^{-1} \B_\mathrm{ISA} \right \|_{\mathrm{Fro}}^2 +\\ & \quad \quad +
   \left \| \W \left( \mathbf{I} - \frac{1}{J} \B_\mathrm{ISA}^\T \B_\mathrm{ISA} \right) \right \|_{\mathrm{Fro}}^2\\ 
    &=  \left \|\frac{1}{\sqrt{J}} \W \B_\mathrm{ISA}^\T - \M_0^\alpha \D^{-1} \right \|_{\mathrm{Fro}}^2 +\\ & \quad \quad +
   \left \| \W \left( \mathbf{I} - \frac{1}{J} \B_\mathrm{ISA}^\T \B_\mathrm{ISA} \right) \right \|_{\mathrm{Fro}}^2,\\
\end{split}
\end{equation}
where the latter term does not depend on $\D_k$ and $\alpha_k^i$ and can be dropped. That yields an equivalent bilinear problem 
\begin{equation}
    \min_{\D,\alpha}
    \left \|\frac{1}{\sqrt{J}} \Delta \W \B_\mathrm{ISA}^\T - \M_0^\alpha \D^{-1} \right \|_{\mathrm{Fro}}^2
\end{equation}
that we minimise by alternating least squares. 

\section{EXPERIMENTS}



\subsection{Torressani's Shark Dataset}

\begin{figure}[b]
\begin{center}
\includegraphics[width=0.75\columnwidth, trim={0.5cm 0.5cm 2cm 1.3cm},clip]{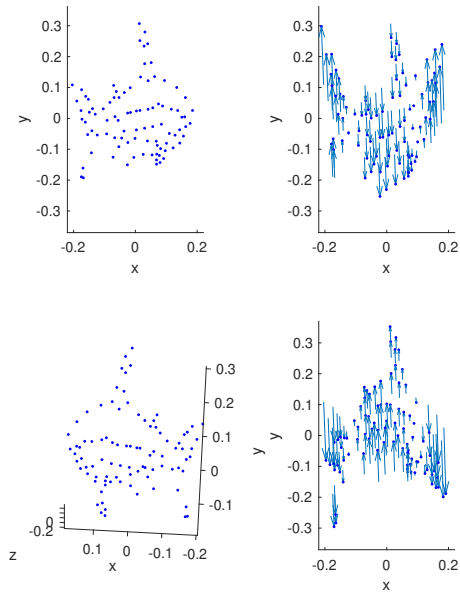}
\end{center}
\vspace{-4mm}
\caption{Affine reconstruction on Torresani's Shark dataset for which the projections onto the image plane can be can be modelled by a single degenerate (planar) 3D shape basis. (Left column) rigid affine 3D Shape from two different directions; (right column) 
illustration of the estimated non-rigid ISA component on both sides around the rigid shape. }\label{fig:shark}
\end{figure}

\begin{table*}[tb]
\caption{Relative reprojection error, reported as the inverse SNR, for the tested NRSFM approaches on different data sets. 
} \label{tab:results}
\centering
\begin{tabular}{|l|c|c|c|c|}
\hline
Inverse SNR $[\%]$ & PI (Dai \etal \cite{Dai12}) & BMM (Dai \etal \cite{Dai12}) & Kong\&Lucey  \cite{Kong16} & Proposed ISA\\
\hline
Shark & 3.5 & 0.33 & 160 & ${ \bf 0.12}^\ddagger$ \\
Balloon & 0.11  & {\bf 0.012} & 1.2 & $0.12^\ddagger$\\
Face LS3D-W & 0.025 & 0.024 & 0.93 & $\bf{0.014}^\ddagger$ \\
Face Binghamton & $\mbox{}^{*}$ & $\mbox{}^{*}$  & 9.3 & $\bf{0.015}^\dagger$ \\
\hline
\end{tabular}\\
{\quad \quad \quad \quad \footnotesize $\mbox{}^\dagger$ by ISA1 variant; \quad $\mbox{}^\ddagger$ by ISA2 variant; \quad
$\mbox{}^{*}$ no result within 24 hours. \hfill}
\vspace{-4mm}
\end{table*}

\begin{figure*}[tb]
\begin{center}
\includegraphics[width=0.85\textwidth, trim={4cm 5cm 3cm 0.5cm},clip]{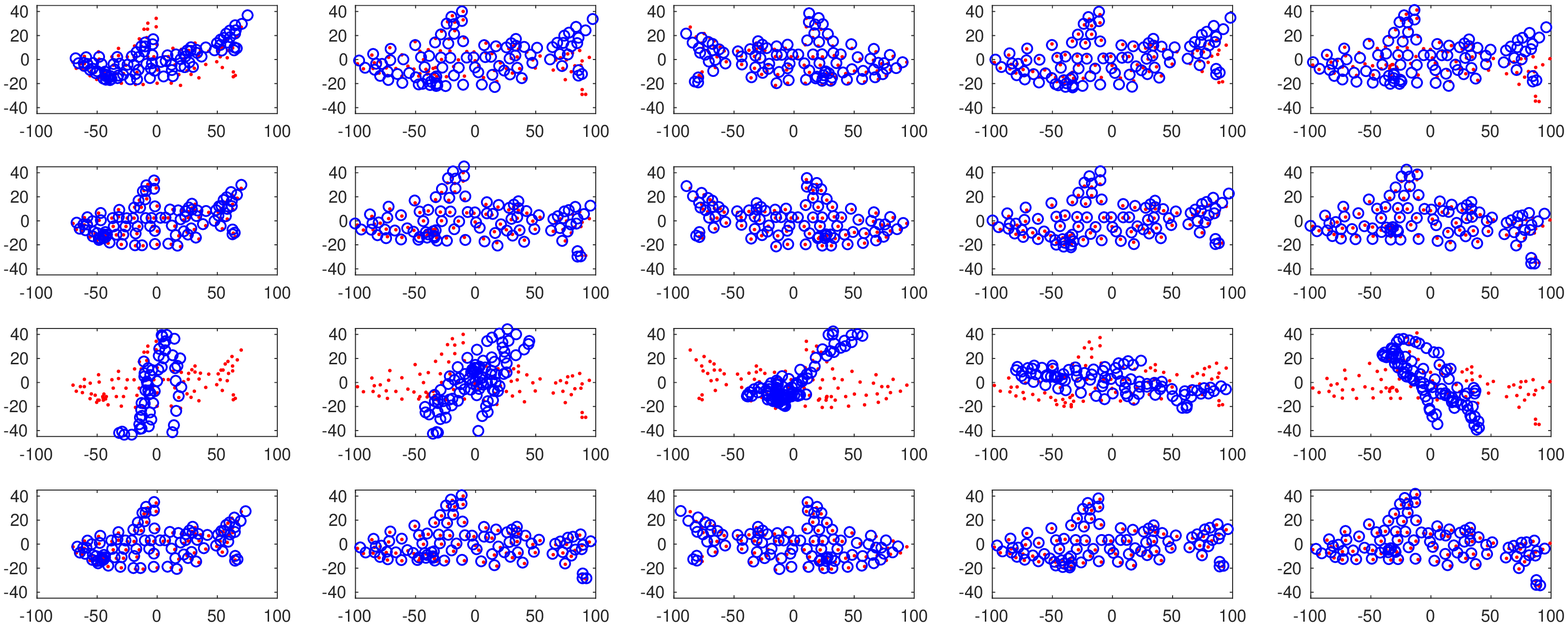}
\end{center}
\caption{Reprojections (blue) to random frames against ground truth projections (red) with the Shark dataset; $K=1$. (1st row) Pseudoinverse Method Dai~\etal ~\cite{Dai12};  (2nd  row) Block Matrix Method Dai~\etal ~\cite{Dai12}; (3rd row) Kong\&Lucey's Priorless Compressible Method~\cite{Kong16}; (4th row) proposed ISA.} \label{fig:sharkReproj}
\vspace{3mm}
\includegraphics[width=\textwidth, trim={4cm 1cm 3cm 0.7cm},clip]{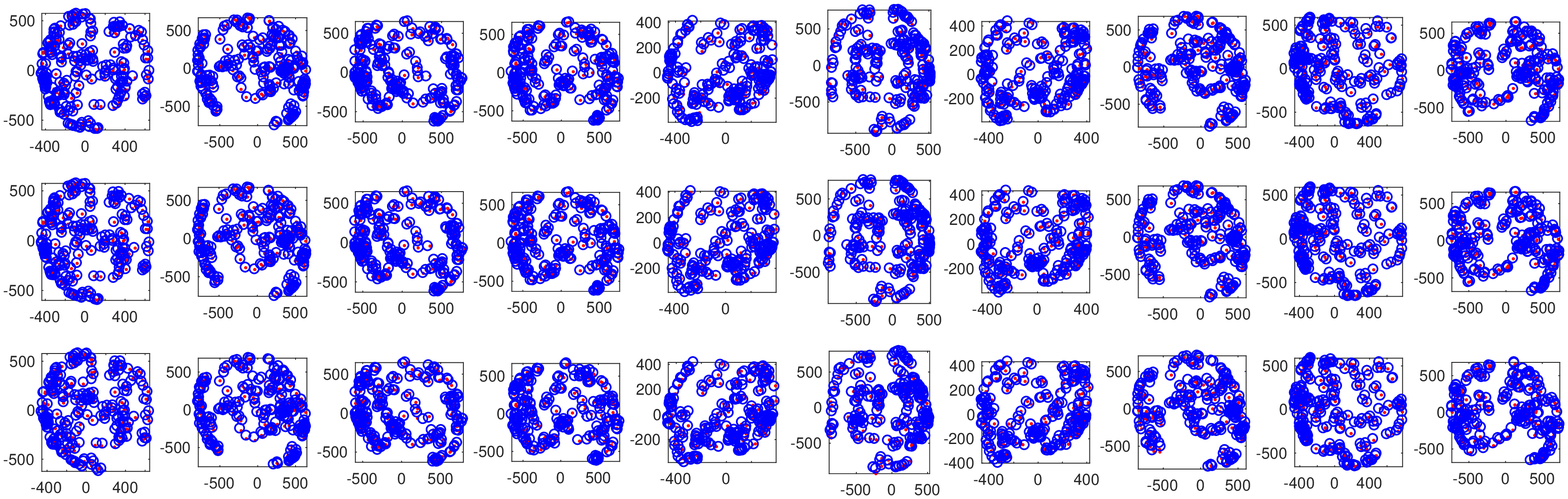}\label{fig:balloonReproj}
\caption{Reprojections (blue) to random frames against ground truth projections (red) with the Balloon deflation dataset; $K=5$. (1st row) Pseudoinverse Method Dai~\etal ~\cite{Dai12};  (2nd  row) Block Matrix Method Dai~\etal ~\cite{Dai12}; (3rd row) Proposed ISA.}
\includegraphics[width=\textwidth, trim={4cm 1cm 3cm 0.7cm},clip]{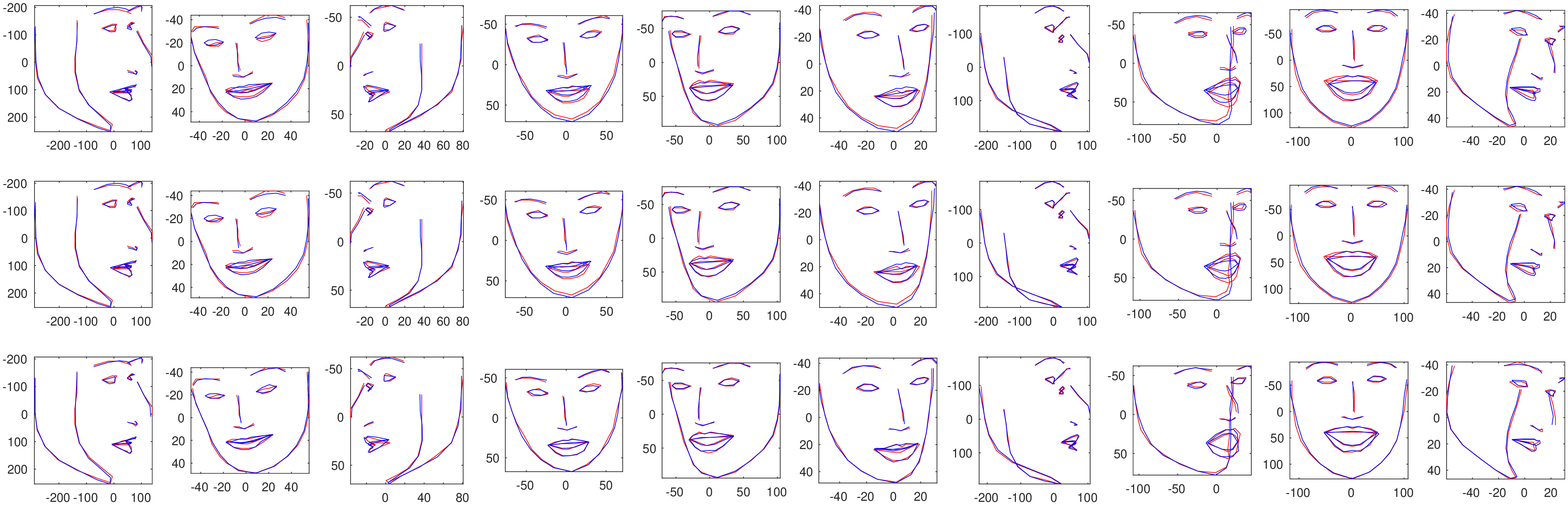}\label{fig:faceReproj}
\caption{Reprojections (blue) to random frames against ground truth projections (red) with the LS3D-W face dataset; $K=9$. (1st row) Pseudoinverse Method Dai~\etal ~\cite{Dai12};  (2nd  row) Block Matrix Method Dai~\etal ~\cite{Dai12}; (3rd row) Proposed ISA.}
\end{figure*}

For the first experiment, we use Torressani's synthetic Shark data set \cite{Torresani08}. It is a degenerate data set with $K=1$. Moreover, the original measurement matrix ($I=240$, $J=91$) has  rank 5 after the translation correction. Hence, the deformation basis is degenerate. This implies a non-unique reconstruction as there will be a 3-parameter-family of solutions for even a single 3D shape basis.
We compared the proposed method against the pseudoinverse method proposed by Dai \etal \cite{Dai12}, as well as their Block Matrix Method, and Kong and Lucey's priorless compressible method 
\cite{Kong16}. Since our method is affine and the reconstruction will be known only up to an unknown affine transformation, it will not be meaningful to compare the results in the 3D space. Instead, we compare the reprojections onto the image plane between the methods. The results are shown in Fig.~\ref{fig:shark}~and~\ref{fig:sharkReproj}, and in Tab.~\ref{tab:results}.

It can be seen from Tab.~\ref{tab:results} that the proposed method (ISA2) achieves the lowest reprojection error, measured by the inverse signal-to-noise-ratio. The other prior free methods do not perform well for this dataset due to the degeneracy, especially, the compressible method failed completely, as also reported in \cite{Kong16}. In spite of the degeneracy of the data set, our method was able to pinpoint the major mode of deformation that is a vector field normal to a reference plane, as Fig.~\ref{fig:shark} illustrates. Here, since only one deformation subspace was considered, the pooling step was trivial.

\subsection{Balloon Deflation}

\begin{figure*}[tb]
\subfigure[]{\includegraphics[width=0.845\textwidth]{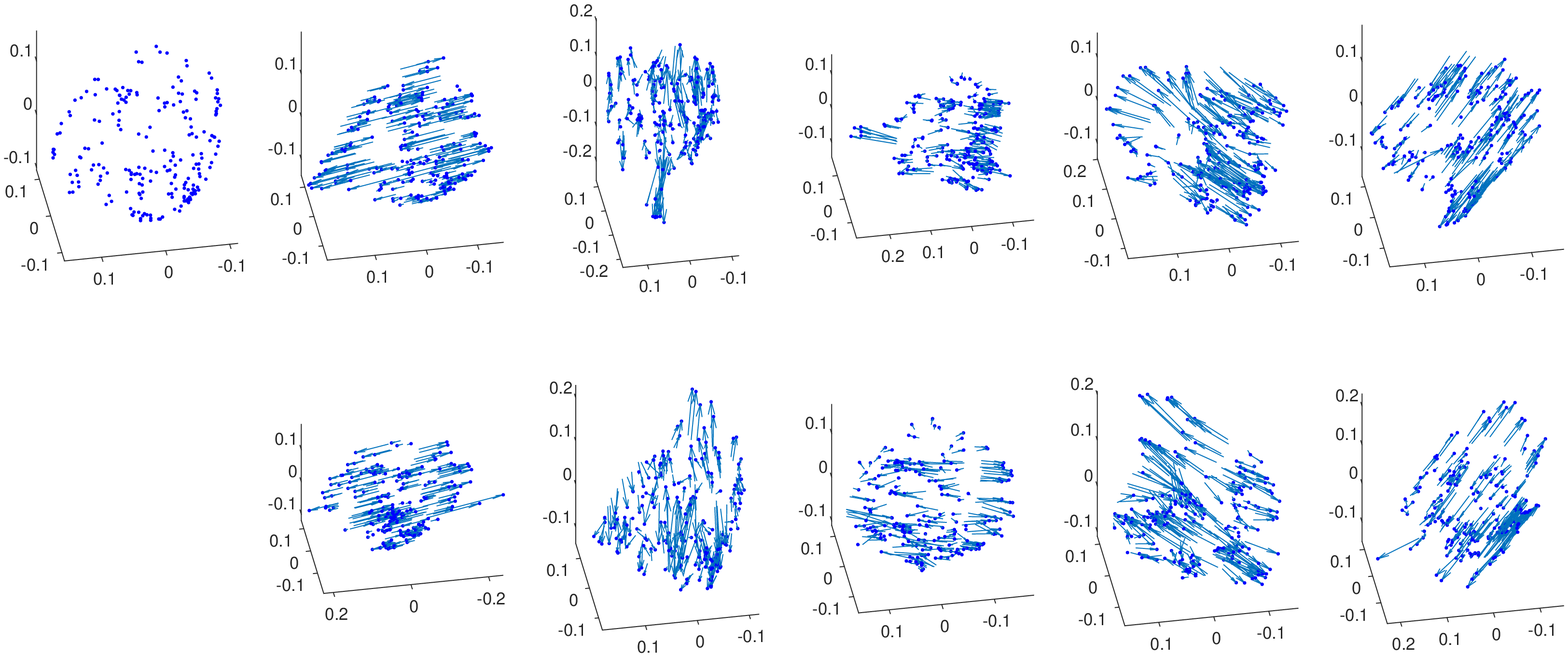}}
\subfigure[]{\includegraphics[width=0.15\textwidth]{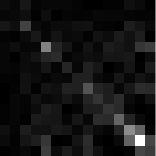}}
\caption{(a) Affine ISA Shape Basis on the Balloon deflation dataset with $K=5$. (left column) Rigid affine 3D Shape $\B_0$; (the other columns) estimated non-rigid ISA shape basis component on both sides around the rigid shape. The arrows illustrate the drift of the points from the mean shape positions. The basis shapes are the components $\B=\B_0 \pm \alpha_k \hat{\B}_k$, where $\alpha_k$ is a positive scalar. (b) The $3K\times3K$ mode covariance matrix $\mathbf{C}$ demonstrating the $3\times3$ block diagonal structure.}\label{fig:balloonModes}
\end{figure*}

For the second test data set, we use the balloon deflation from the NRSfM Challenge 2017 \cite{Jensen18}. It is a simulated data set with $I=51$ projections generated by reprojecting real tracked 3D data points ($J=211$) by a virtual, perspective camera having a circular camera trajectory. By using an affine camera model, we can thus only achieve an approximation of the ground truth camera geometry. We then estimated the result using the reference approaches 
and our ISA methods. 
We assumed five deformation modes $(K=5)$. The results are in Tab.~\ref{tab:results}, and Fig.~\ref{fig:balloonReproj}~and~\ref{fig:balloonModes}. 

For this data set, the Block Matrix Method of~\cite{Dai12} gave the best result, whereas the Pseudoinverse and the proposed ISA approach obtain similar scores. Fig.~\ref{fig:balloonModes} illustrates the estimated, statistically independent 
modes. The third component most clearly indicates the size change, whereas the other modes represent different kinds of non-linear shape deformations.
The mode covariance matrix (Fig.~\ref{fig:balloonModes}b) shows that the highest correlations are concentrated onto the diagonal, hence, the independence assumption is fair. Nonetheless, there are some off-block-diagonal-correlations that most likely contributed to the higher score. 

\subsection{Face LS3D-W Dataset}

\begin{figure*}
\subfigure[]{\includegraphics[width=0.85\textwidth, trim={7cm 1cm 5cm 0.5cm},clip]{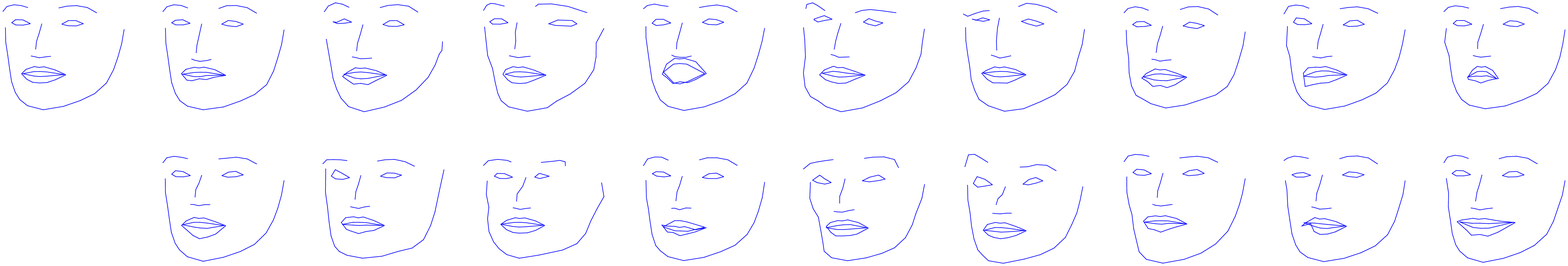}}%
\subfigure[]{\includegraphics[width=0.15\textwidth]{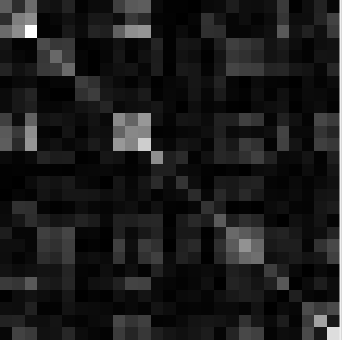}}
\caption{(a) Affine ISA shape basis for the LS3D-W data set with $K=9$. (Left column) rigid affine 3D Shape $\B_0$; (the other columns) the 9 estimated 3D ISA basis shapes $\B=\B_0 \pm \alpha_k \hat{\B}_k$, where $\alpha_k$ is a positive scalar. (b) The $3K\times3K$ mode covariance matrix $\mathbf{C}$ demonstrating the $3\times3$ block diagonal structure. 
}\label{fig:faceModes}
\end{figure*}

For the third experiment, we use the LS3D-W data set \cite{Bulat2017} consisting of matched feature points for 7200 human faces with various expressions. Each face contains 68 2D feature points that were automatically found and matched, as described in \cite{Bulat2017}. The faces were in random orientation and order so no temporal smoothness could be applied. We compare our method (ISA2) against Dai's \cite{Dai12} and Kong and Lucey's \cite{Kong16} methods. Dai's method is computationally most demanding due to the size of the database: the computation of the result took about 2 CPU days, Kong and Lucey's about 6 CPU hours. In contrast, an ISA estimate could be computed in about twenty CPU minutes. The results are shown in Tab.~\ref{tab:results}, and in Figs.~\ref{fig:faceReproj}~and~\ref{fig:faceModes}.

From the results, it can be seen that the proposed method gave the best numerical results with almost a half of the inverse SNR when compared to either of the methods by Dai \etal \cite{Dai12}. When looking at the reprojections, it can be seen that their approach had more difficulties in reproducing the fine structure of the mouth (see columns 2, 4, 6, 8, and 9 in Fig.~\ref{fig:faceReproj}) than the proposed method. Each estimated basis shape, shown in Fig.~\ref{fig:faceModes}a, demonstrate a clear semantic interpretation. From the mode correlation matrix (Fig.~\ref{fig:faceModes}b) it can be seen that the strongest off-block-diagonal covariance is between the the first and fourth basis shape. One can also note that the lips are slightly distorted in both modes that suggest that there is in fact a statistical dependency between the modes while the statistical independence assumption yields an accurate approximation for the shapes and poses in the data set.

\subsection{Binghamton 3D Facial Expression Dataset}
\begin{figure*}
    \begin{center}
    \subfigure[]{\includegraphics[width=0.18\textwidth]{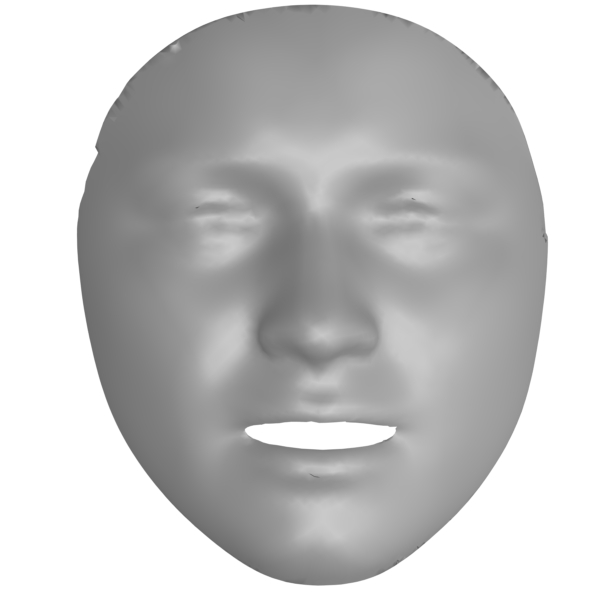}}
    \subfigure[]{\includegraphics[width=0.18\textwidth]{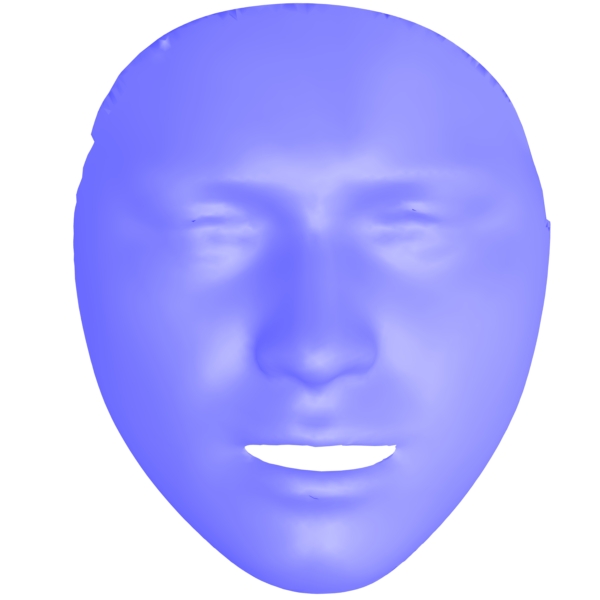}}
    \subfigure[]{\includegraphics[width=0.18\textwidth]{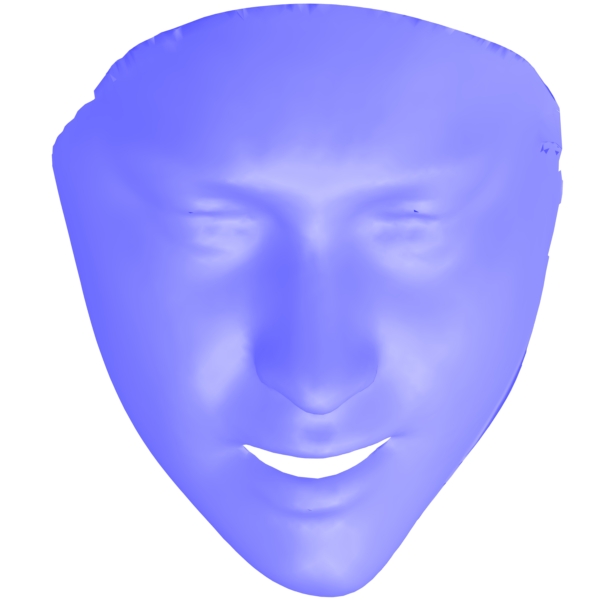}}
    \subfigure[]{\includegraphics[width=0.18\textwidth]{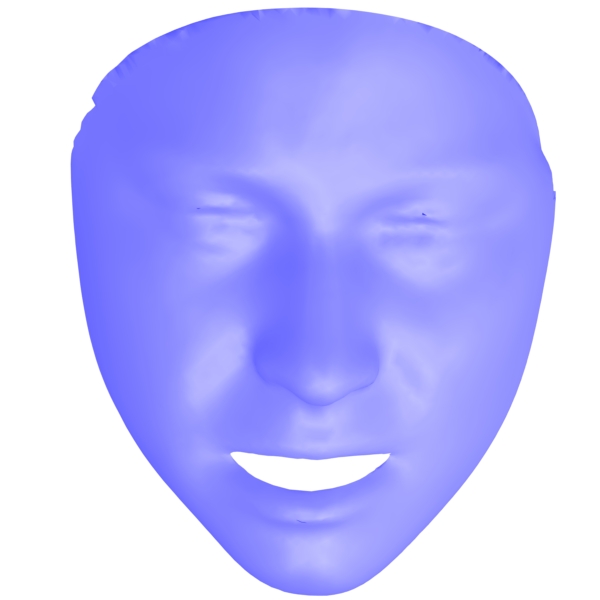}}
    \subfigure[]{\includegraphics[width=0.18\textwidth]{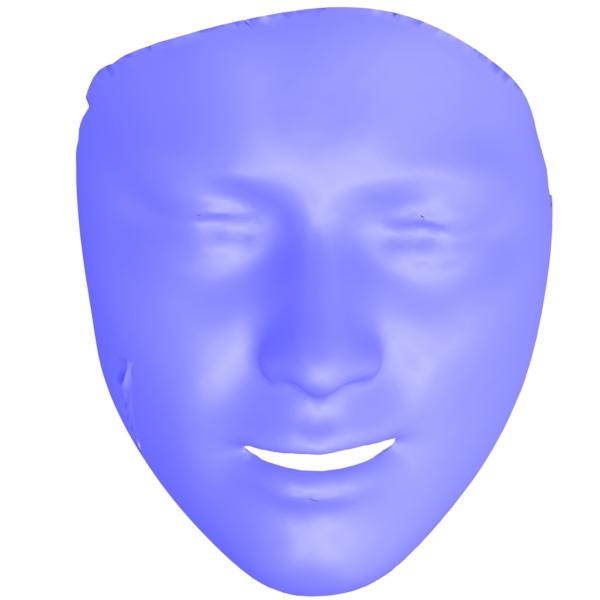}}
    \includegraphics[width=0.18\textwidth]{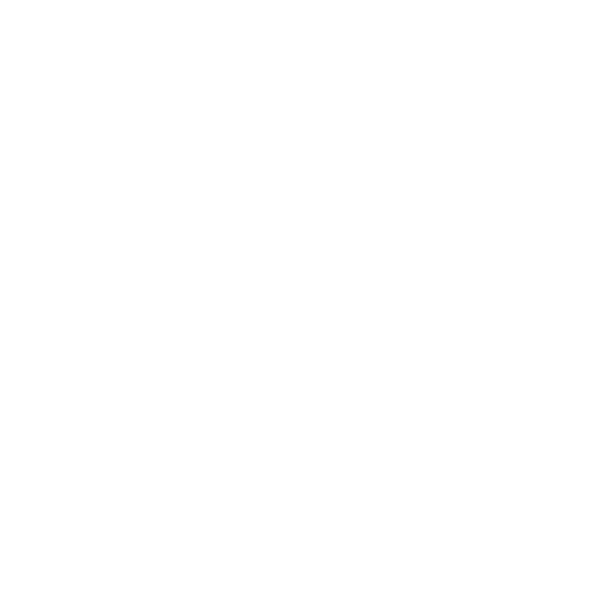}
    \subfigure[]{\includegraphics[width=0.18\textwidth]{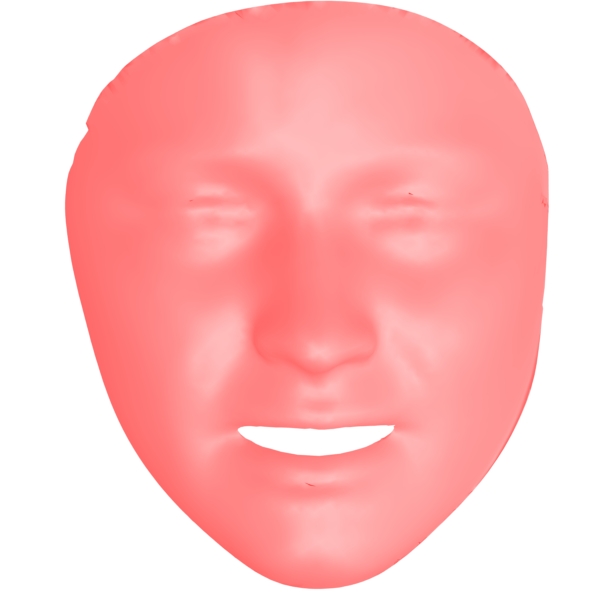}}
    \subfigure[]{\includegraphics[width=0.18\textwidth]{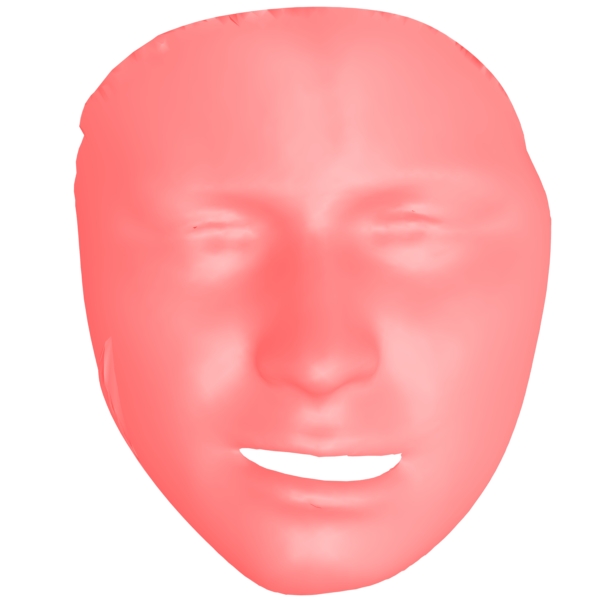}}
    \subfigure[]{\includegraphics[width=0.18\textwidth]{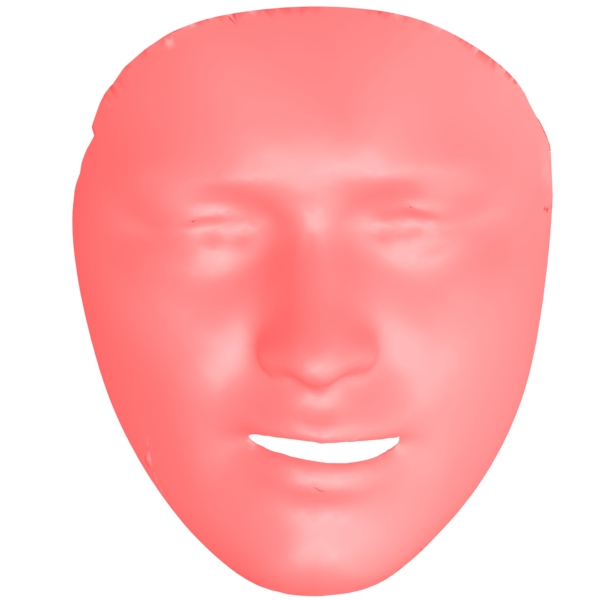}}
    \subfigure[]{\includegraphics[width=0.18\textwidth]{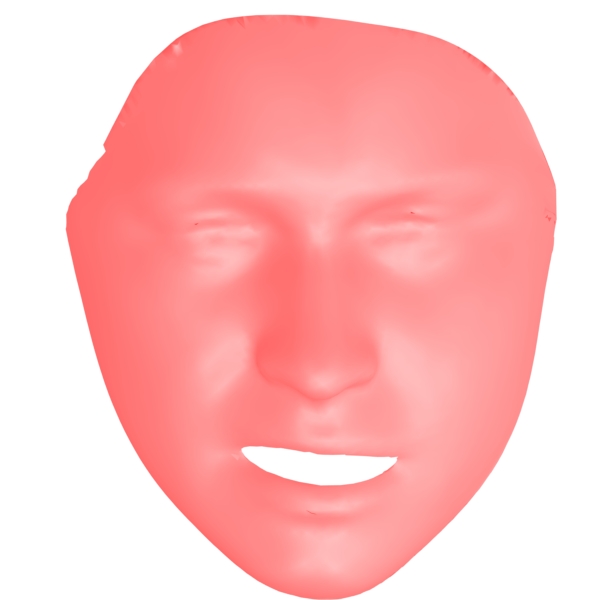}}
    \includegraphics[width=0.18\textwidth]{./bu3dfe/K_09/white_dummy.jpg}
    \subfigure[]{\includegraphics[width=0.18\textwidth]{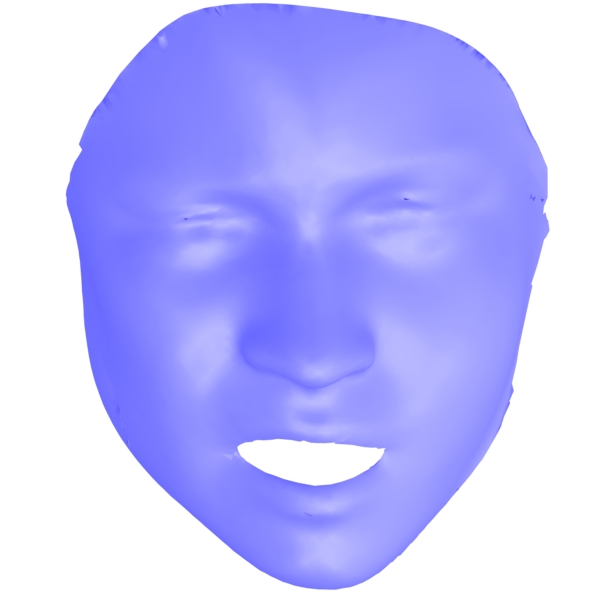}}
    \subfigure[]{\includegraphics[width=0.18\textwidth]{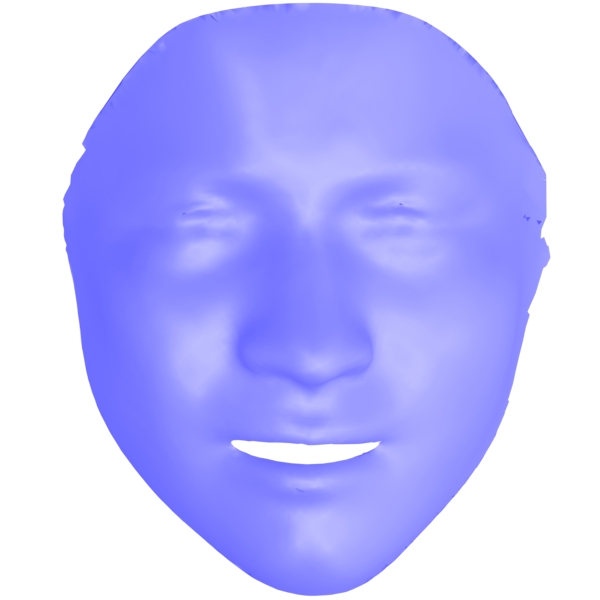}}
    \subfigure[]{\includegraphics[width=0.18\textwidth]{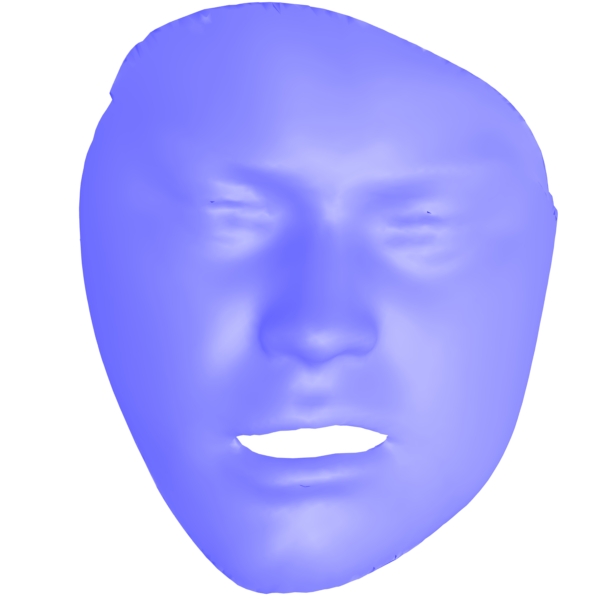}}
    \subfigure[]{\includegraphics[width=0.18\textwidth]{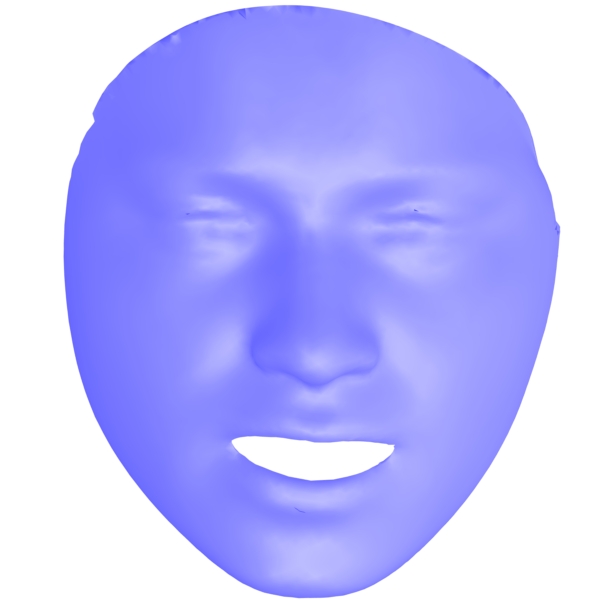}}
    \subfigure[]{\includegraphics[width=0.18\textwidth]{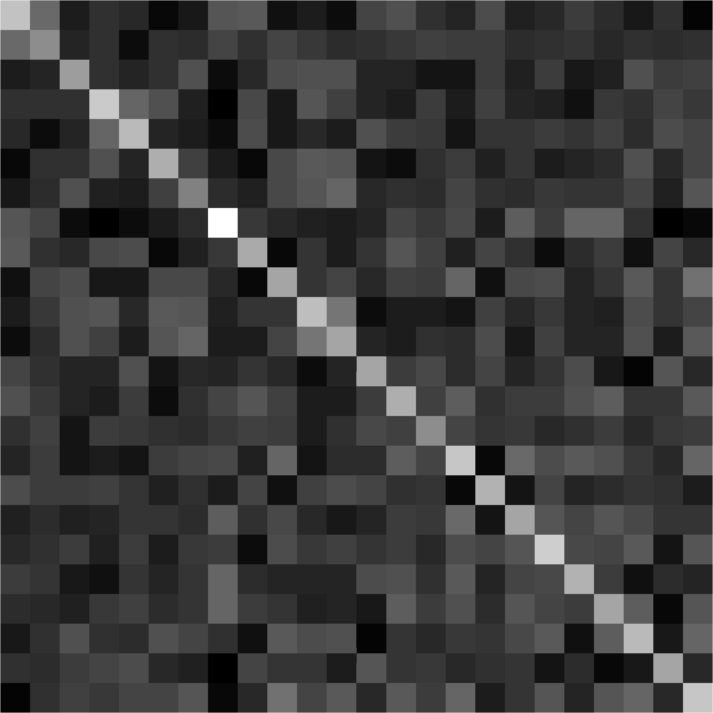}}
    \subfigure[]{\includegraphics[width=0.18\textwidth]{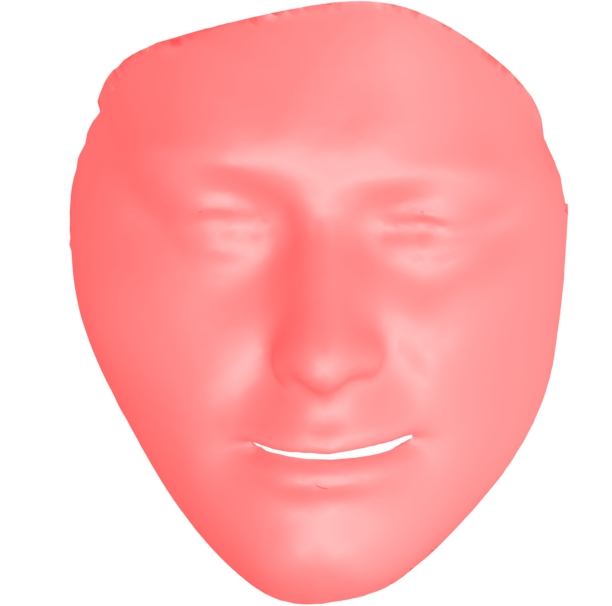}}
    \subfigure[]{\includegraphics[width=0.18\textwidth]{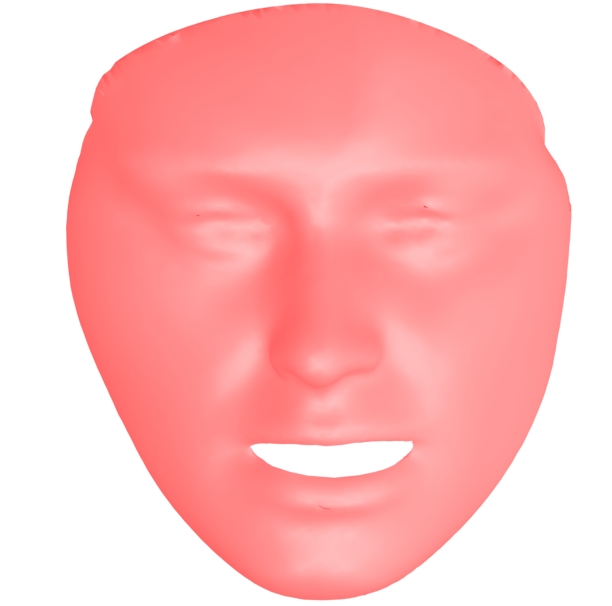}}
    \subfigure[]{\includegraphics[width=0.18\textwidth]{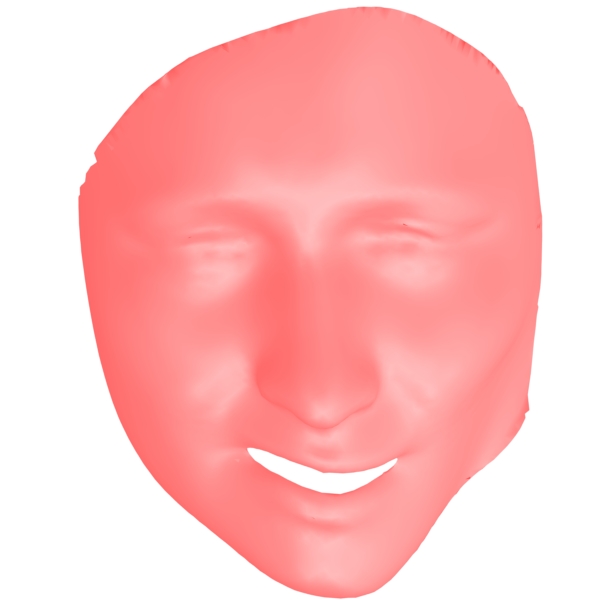}}
    \subfigure[]{\includegraphics[width=0.18\textwidth]{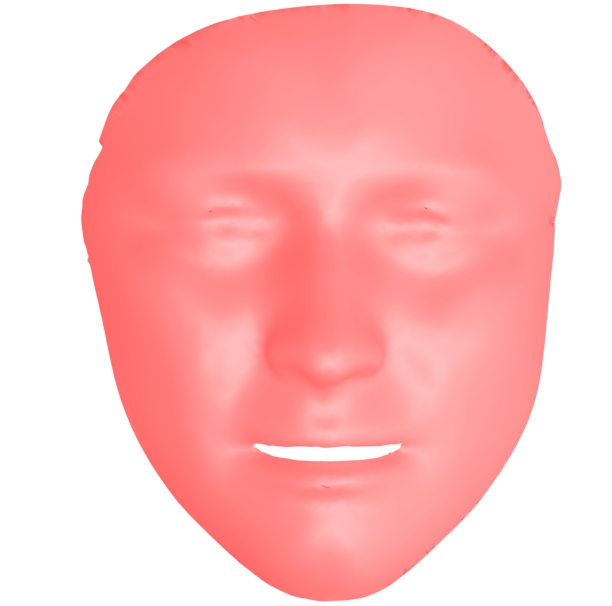}}
    \end{center}
    \caption{Reconstructions of the $K=8$ basis shapes computed from the Binghamton, BU3DFE~\cite{BU3DFE} dataset. 
    (a) The mean rigid shape $\B_0$; (b--r, except n) the estimated 3D ISA basis shapes $\B=\B_0 \pm \alpha_k \hat{\B}_k$, shown in red and blue, respectively. (n) Mode covariance matrix.} \vspace{-3mm}
    \label{fig:bu3dfe}
\end{figure*}

The data set~\cite{BU3DFE} contains $25$ shapes of $100$ subjects with $7$ different expressions (\emph{neutral}, \emph{happy}, \emph{disgusted}, \emph{fear}, \emph{angry}, \emph{surprised}, \emph{sad}) recorded by a 3D-face scanner. All the expressions, except the neutral, were recorded in four different strengths. The subjects had varying ethnic background and their age range was from 18 to 70 years. A total of $56\%$ of the subjects were female and $44\%$ male. We obtained $7308$ 3D-correspondences between the shapes by non-rigid registration~\cite{Golyanik2016:ECPD}. 2D correspondences, simulated from the 3D correspondences, were used as the input for the experiment. 
%
Results are shown in Tab.~\ref{tab:results}, and Fig.~\ref{fig:bu3dfe}.  A comparison with the baseline algorithms by Dai \etal~\cite{Dai12} 
was not possible since both methods did not converge within a reasonable amount of time.
The result by Kong and Lucey~\cite{Kong16} was modest probably due to the fact there was no temporal structure in the data. 
Our method (ISA1) was able to produce an accurate fit, as the low inverse SNR demonstrates. Also, as can be seen in Fig.~\ref{fig:bu3dfe}, the estimated shape basis was able to capture the major structure variations in the database, including those  related to person expression changes. From the mode covariance marix (Fig.~\ref{fig:bu3dfe}n), it can be seen that the covariance was concentrated onto the diagonal while the statistical dependences are not as strong as with the LS3D-W data set. 

\section{CONCLUSIONS}

In this paper we proposed a generalisation for non-rigid structure-from-motion. In contrast to the earlier belief that the recovery of shape basis would be ambiguous without prior information, we have shown that only assuming statistical independence between the 3D basis shapes yields an uncalibrated, affine shape basis and affine non-rigid structure and motion estimates. In analogy to the theory about rigid structure-from-motion, 
estimating an affine reconstruction instead of an Euclidean one yields a simpler solution for the non-rigid structure-from-motion problem, and independent subspace analysis serves as a natural tool to resolve the basis ambiguity. Our experiments showed that the approach is suitable for large data sets and it facilitates modelling and analysis of non-rigid structures in an uncalibrated setting. The approach hence opens the way for solving the non-rigid structure-from-motion problem. In future, we are going to extend our methodology to handle missing data and to cope with more versatile statistical dependencies between the shape bases. 

\appendix

\section{IRLS FOR BLOCK-FORM RECOVERY} \label{app:IRLS}

\noindent \emph{Problem:} Find $\alpha_k^i \in \mathbb{R}$ and $\D_k \in \mathbb{R}_{3 \times 3}$ such that 
\begin{equation}
    \sum_{i,k} \| \M_k^i \D_k - \alpha_k^i \M^i \|^2_\mathrm{Fro} \longrightarrow \min,  \label{eq:minproblem}
\end{equation}
subject to
\begin{equation}
    \| \D_k \|_\mathrm{Fro} = 1, \ k=1,2,\ldots,K,
\end{equation}
where $\M_k^i,\M^i \in \mathbb{R}_{2 \times 3}$. 

\vspace{2mm}
\noindent \emph{Solution:} 

Let $\mathbf{d}_k=\mathrm{vec} (\D_k)$,  $\mathbf{m}^i = \mathrm{vec} (\M^i)$. Now, for $i=1,2,\ldots,I$, $k=1,2,\ldots,K$,
\begin{equation}
\begin{split}
    \| \M_k^i \D_k - & \alpha_k^i \M^i \|_\mathrm{Fro}^2 \\
    &= \Bigg \| 
\underbrace{\begin{pmatrix}
    \M_k^i & \mathbf{0} & \mathbf{0} \\
     \mathbf{0} & \M_k^i &  \mathbf{0} \\
      \mathbf{0} & \mathbf{0} & \M_k^i \\
\end{pmatrix}}_{\triangleq \mathbf{N}_k^i} 
\mathbf{d}_k - \alpha_k^i \mathbf{m}^i
\Bigg \|_2^2\\
    &= \left \| \begin{pmatrix}\mathbf{N}_k^i & -\mathbf{m}^i \end{pmatrix} 
    \begin{pmatrix}
        \mathbf{d}_k\\
        \alpha_k^i
    \end{pmatrix}
    \right \|_2^2.
\end{split}
\end{equation}
By collecting all the coefficients $\mathbf{N}_k^i,\mathbf{m}^i$ into a $6IK \times (9+I)K$ matrix $\mathbf{N}$
the problem \eqref{eq:minproblem} is equivalent to the constrained least squares problem 
\begin{equation}
    \begin{split}
    \|  \mathbf{N} 
     ( \mathbf{d}_1,
        \ldots,
        & \mathbf{d}_K,
        \alpha_1^1,
         \ldots,
        \alpha_K^1,
        \alpha_1^2,
         \ldots,
        \alpha_K^2,
         \ldots,
        \alpha_K^I) \|_2 \\ &\longrightarrow \min,
    \end{split} \label{eq:modminproblem} \vspace{-2mm}
\end{equation}
subject to $||\mathbf{d}_k\|_2=1$, $k=1,2,\ldots,K$. 
The estimate 
can be found by the iteratively reweighted least squares by first assuming that $\alpha_k^{1,(n)}=1/K$ for $n=0$, $k=1,\ldots,K$ and finding the solution of the reduced system
\begin{equation}
    \begin{split}
    \|  \mathbf{N}_{\backslash \alpha^1} 
     ( \mathbf{d}_1,
        \ldots,
        & \mathbf{d}_K,
        \alpha_1^2,
         \ldots,
        \alpha_K^2,
         \ldots,
        \alpha_K^I) - \mathbf{c}^{(n)} \|_2 \\ &\longrightarrow \min,
    \end{split} 
\end{equation}    
where ${\mathbf{N}}_{\backslash \alpha^1}$ is constructed from $\mathbf{N}$ by dropping the columns corresponding to $\alpha_k^1$, for all $k$, and $\mathbf{c}^{(n)}= -{\mathbf{N}}_{\alpha^1} (\alpha_1^{1,(n)},\ldots,\alpha_K^{1,(n)})$. The estimate for $\alpha_k^{1,(n+1)}$ is computed as $\alpha_k^{1,(n+1)} \leftarrow  \alpha_k^{1,(n)} / \| \mathbf{d}_k^{(n)} \|$, and the computation is iterated until convergence. This reweighting follows from the weighting $w_k^i= \| \mathbf{d}_k^{(n)} \|^{-2}$ in the iterated reweighted least squares (IRLS) scheme seeking to adjust the mixing weights in the first view that results in the unity Frobenius norms for the estimate of $\mathbf{D}_k$. The IRLS solution typically converges in only a few iterations, so the computational overhead is negligible.  

%


\bibliographystyle{ieee}

\bibliography{references}

%
%
%
%

\end{document}